\documentclass{article}

\usepackage{arxiv}

\usepackage{amsmath,amsfonts,amssymb}
\usepackage{array}
\usepackage{booktabs}
\usepackage{pdfpages}
\usepackage{multirow}
\usepackage{tabularx}
\usepackage{graphicx}
\usepackage[caption=false,font=normalsize,labelfont=sf,textfont=sf]{subfig}
\usepackage{textcomp}
\usepackage{stfloats}
\usepackage{url}
\usepackage{verbatim}
\usepackage{cite}
\usepackage{algorithm}
\usepackage{orcidlink}
\usepackage{algpseudocode}
\usepackage{pgfplots}
\usepackage{wrapfig}
\usepackage{pgfplotstable}
\usepackage{graphicx}
\usepackage{multirow}
\usepackage{amsmath,amssymb,amsfonts}
\usepackage{amsthm}
\usepackage{mathrsfs}
\usepackage[title]{appendix}
\usepackage{xcolor}
\usepackage{textcomp}
\usepackage{manyfoot}
\usepackage{booktabs}
\usepackage{algorithm}
\usepackage{algorithmicx}
\usepackage{algpseudocode}
\usepackage{listings}
\usepackage[utf8]{inputenc}
\usepackage[T1]{fontenc}
\usepackage{hyperref}
\usepackage{orcidlink}
\usepackage{array}
\usepackage{tabularx}
\usepackage{url}
\usepackage{enumitem}
\usepackage{pgfplots}
\usepackage{tikz}
\usepackage[T1]{fontenc}
\usepackage{mathptmx}
\usepackage{graphicx}
\usepackage{amsmath,amssymb}
\usepackage{booktabs}
\usepackage{multirow}
\usepackage{array}
\usepackage{tabularx}
\usepackage{enumitem}
\usepackage{float}
\usepackage{url}
\usepackage{cite}
\usepackage{caption}
\usepackage{microtype}
\newcommand{\ba}{\mathrm{BA}}

\pgfplotsset{compat=1.18}
\usetikzlibrary{arrows.meta,positioning,calc}

\definecolor{designBlue}{HTML}{2563EB}
\definecolor{methodPurple}{HTML}{7C3AED}
\definecolor{conditionOrange}{HTML}{EA580C}
\definecolor{evidenceGreen}{HTML}{15803D}
\definecolor{provTeal}{HTML}{0F766E}
\definecolor{auditRed}{HTML}{B91C1C}
\definecolor{analysisIndigo}{HTML}{4338CA}
\definecolor{arrowDark}{HTML}{374151}
\definecolor{textDark}{HTML}{111827}

\title{\textbf{When Adaptation Hurts}:\\
Split Sensitivity and Person-Level Negative Transfer in Federated Wearable Onboarding}

\author{
\textbf{Rahil Aftab}~\orcidlink{0009-0009-0983-4726}\\
Department of Computer Science and Engineering\\
Jamia Hamdard\\
New Delhi 110062, India\\
\texttt{rahilaftab12@gmail.com}
\and
\textbf{Vineet Kumar Rakesh}~\orcidlink{0009-0000-7102-6564}\\
Engineering Science\\
Homi Bhabha National Institute\\
Anushaktinagar, Mumbai 400094, Maharashtra, India\\
Computer and Informatics Group\\
Variable Energy Cyclotron Centre\\
1/AF, Bidhannagar, Kolkata 700064, West Bengal, India\\
\texttt{vineet@vecc.gov.in}
\and
\textbf{Soumya Mazumdar}~\orcidlink{0009-0006-3521-9557}\\
Department of Computer Science and Business Systems\\
Gargi Memorial Institute of Technology\\
Affiliated to Maulana Abul Kalam Azad University of Technology\\
Balarampur, Mouza Beralia, Baruipur, Kolkata 700144, West Bengal, India\\
\texttt{reachme@soumyamazumdar.com}
\and
\textbf{Tapas Samanta}~\orcidlink{0000-0003-0521-0747}\\
Engineering Science\\
Homi Bhabha National Institute\\
Anushaktinagar, Mumbai 400094, Maharashtra, India\\
Computer and Informatics Group\\
Variable Energy Cyclotron Centre\\
1/AF, Bidhannagar, Kolkata 700064, West Bengal, India\\
\texttt{tsamanta@vecc.gov.in}
}

\begin{document}
\maketitle

\begin{abstract}
Federated wearable models eventually serve people absent from source training, but favorable average accuracy does not establish that unlabeled onboarding helps each person. We evaluate six core onboarding strategies on five wearable datasets under a leakage-controlled protocol that fixes source checkpoints, estimates normalization from source data only, separates calibration from evaluation recordings, and performs inference over held-out people rather than windows, devices, or random seeds. Completing all eligible HHAR and PAMAP2 outer-person rotations materially changes the conclusion obtained from the original frozen fold. On HHAR, balanced accuracy on that single person is 95.6--97.2\% across methods versus 78.3--83.0\% over all nine users, a reduction of 13.8--17.8 percentage points (pp). The displayed mean leader changes on both datasets, while paired leader--runner bootstrap intervals include zero and do not resolve a superior method. No adaptive core mechanism combines positive mean gain in all five datasets with zero seed-averaged person-level losses greater than 2 percentage points (pp). FedBN has one such loss and ATP-style adaptation has eight; Feature-only has none after seed averaging, but its exact one-sided 95\% upper bound is 7.6\%. A complementary seed--person stress audit records 4, 22, and 10 harmful realizations out of 114 for FedBN, ATP-style, and Feature-only, respectively; these are repeated realizations, not independent participants. Tail quality, calibration availability, and fall-window specificity reveal additional failures hidden by mean accuracy. The study therefore provides an auditable development benchmark and failure map rather than a universal-superiority or deployment-safety claim.
\end{abstract}

\textbf{Keywords:}
federated learning; wearable sensing; human activity recognition; test-time adaptation; personalization; negative transfer

\section{Introduction}
\label{sec:introduction}

Federated learning (FL) trains models from decentralized client data without
collecting those raw records at a central server \cite{mcmahan2017}. Deployment
creates a second problem: the trained model may later serve a person who never
participated in the source federation. For wearable sensing, this unseen client
can differ in physiology, movement style, sensor placement, device hardware,
recording session, and activity prevalence. The resulting shift can be large
even when the model architecture and label space remain unchanged.

A short calibration stream is often realistic at onboarding. Because manual
labels are expensive or unavailable, test-time adaptation (TTA) and federated
test-time personalization seek to adapt from unlabeled target observations
\cite{wang2021tent,bao2023atp}. This is attractive but not intrinsically safe.
Small or class-incomplete calibration sets can drive entropy minimization or
normalization toward a worse predictor; even a positive dataset mean can hide
negative transfer for particular people. Existing TTA benchmarks already show
that conclusions depend on source-model quality, shift type, model selection,
batch size, and device constraints
\cite{zhao2023pitfalls,zhao2024nanoadapt,danilowski2026botta}. The unresolved
question is whether an apparently successful federated wearable onboarding
result survives recording separation and inference across all eligible people.

Wearable evaluation makes these risks easy to understate. Overlapping physical
episodes can leak across partitions, device endpoints from one participant can
be counted as independent samples, and repeated random seeds can be mistaken
for additional people. Window-level confidence intervals then become
artificially narrow. Fall-oriented datasets add asymmetric consequences: a
model can attain high fall sensitivity while producing many false-positive
windows. Without clustering correlated windows into episodes, these counts do
not yet estimate operational false alarms. Mean window accuracy alone therefore
does not answer the deployment question.

We study a narrower and operationally testable question:
\begin{quote}
\emph{When does unlabeled onboarding improve an unseen federated wearable
client, when does it cause negative transfer, and which evaluation controls are
needed to reveal that harm?}
\end{quote}

\begin{enumerate}
    \item[\textbf{RQ1.}] \textbf{Persistence across outer-person rotations:}
    Does the conclusion obtained from a single held-out person persist after
    completing all eligible outer-person rotations?

    \item[\textbf{RQ2.}] \textbf{Adaptive utility--harm trade-off:}
    Can an adaptive mechanism simultaneously achieve strong cross-domain mean
    utility and low person-level harm?

    \item[\textbf{RQ3.}] \textbf{Calibration and tail behavior:}
    Do calibration quantity and mean quality adequately characterize
    tail behavior?

    \item[\textbf{RQ4.}] \textbf{Fall sensitivity and specificity:}
    What changes when fall sensitivity and specificity are evaluated jointly?
\end{enumerate}

The paper makes four contributions:
\begin{enumerate}[leftmargin=*]
    \item An onboarding validity contract that combines federated source
    provenance, immutable participant splits, recording-disjoint unlabeled
    calibration/evaluation, outer-person inference, matched zero-shot branches,
    and explicit calibration availability.
    \item A frozen two-layer benchmark that retains a complete 21-method broad
    screen while giving six scientifically distinct core strategies complete
    nine-user HHAR and eight-subject PAMAP2 rotations.
    \item Direct evidence of split and difficulty sensitivity: the original
    frozen HHAR person's BA exceeds every corresponding complete-rotation mean
    by 13.8--17.8 pp, while paired uncertainty shows that the displayed
    leader changes are unresolved near-ties rather than established superiority.
    \item A two-estimand failure analysis that separates seed-averaged
    participant effects from descriptive seed--person deployment realizations,
    adds finite-sample bounds for zero observed harm, and retains tail quality,
    availability, cost, and fall specificity rather than reducing onboarding to
    one average score.
\end{enumerate}

This is a retrospective, audited development benchmark rather than an
independent confirmatory superiority trial. Earlier candidate and selector
criteria were not met, all unfavorable results remain visible, and no method is
promoted as a new universal winner. This status limits superiority and safety
claims, but it does not invalidate the predeclared rotation extension or the
protocol failures that extension reveals.

\section{Related Work}
\label{sec:related}

\subsection{Federated heterogeneity and personalization}

FedAvg aggregates local updates into a global model \cite{mcmahan2017}, while
FedProx modifies local objectives to accommodate heterogeneous clients
\cite{li2020fedprox}. Personalized FL instead permits client-specific models or
components. FedPer retains personal classifier layers \cite{arivazhagan2019fedper};
FedBN keeps normalization local under feature shift \cite{li2021fedbn};
Per-FedAvg meta-learns an initialization for client adaptation
\cite{fallah2020perfedavg}; and Ditto optimizes related global and personal
models \cite{li2021ditto}. FedL2P learns client-dependent personalization
strategies and argues against a single hand-crafted strategy for every client
\cite{lee2023fedl2p}.

Most personalization methods assume that the target client participates in
federated training or possesses labeled local data. Out-of-federation
generalization instead asks how a source federation serves unseen clients;
topology-aware FL is one recent training-time approach to that problem
\cite{ma2024beyond}. Our setting keeps source training fixed and asks whether a
previously unseen client should adapt from a short \emph{unlabeled} stream.

\subsection{Test-time adaptation and federated onboarding}

TENT adapts normalization parameters by prediction-entropy minimization
\cite{wang2021tent}; EATA filters unreliable or redundant samples and
regularizes important weights to reduce forgetting \cite{niu2022eata}. ATP
formalizes adaptive test-time personalization in FL and learns module-wise
adaptation rates from source-domain shifts \cite{bao2023atp}. These methods
motivate label-free onboarding, but their original evaluations and backbones
do not resolve wearable-specific participant and recording boundaries.

TTAB identifies sensitivity to model selection, source-model quality, and
shift type \cite{zhao2023pitfalls}. NanoAdapt demonstrates negative transfer
under extremely small TTA batches \cite{zhao2024nanoadapt}, while continual-TTA
work highlights error accumulation and unstable online updating under changing
or small-batch conditions \cite{wang2022cotta,niu2023sar}. BoTTA evaluates TTA
under mobile and edge constraints \cite{danilowski2026botta}. These studies
establish that TTA evaluation itself is a research problem; consequently, our
novelty claim is not that negative transfer exists. We instead study its
person-level frequency and severity when the source model was trained
federatively and calibration recordings are unavailable to final evaluation.
Implementations named \emph{port} or \emph{proxy} below are in-repository
research references, not author-validated reproductions.

\subsection{Federated wearable recognition and evaluation integrity}

Federated transfer and semi-supervised personalization have been investigated
for wearable health and activity recognition \cite{chen2020fedhealth,presotto2022fedar}.
Cross-person wearable TTA now includes normalization/prototype adaptation in
OFTTA \cite{wang2023oftta}, contrastive online adaptation in COA-HAR
\cite{rey2026coahar}, and supervised/unsupervised prototype updates under
limited user calibration \cite{burzer2026uncertainty}. Recent preprints further
specialize temporal routing and physics-informed constraints for wearable
streams \cite{zhou2026sight,li2026pitta}. These works establish that unseen-user
and unlabeled wearable adaptation are not new by themselves.
OFTTA, for example, reports each target domain but adapts online on the test
stream; our protocol instead freezes a bounded calibration pool from separate
recordings and evaluates on untouched recordings. Our source is additionally a
federation rather than a pooled empirical-risk model.

Public datasets expose complementary shifts: WISDM \cite{kwapisz2011wisdm},
PAMAP2 \cite{reiss2012pamap2}, HHAR \cite{stisen2015hhar}, SisFall
\cite{sucerquia2017sisfall}, UniMiB SHAR \cite{micucci2017unimib}, and UP-Fall
\cite{martinez2019upfall}. Yet sliding-window overlap and non-subject-independent
evaluation can materially inflate HAR results \cite{dehghani2019windows}.
The gap addressed here is therefore not another favorable mean or a claim to
be the first wearable TTA method. It is the conjunction of federated source
training, recording-disjoint unlabeled onboarding, complete outer-person
inference where feasible, branch-matched harm, tail quality, calibration
availability, fall false-positive windows, and artifact-level provenance. Representative
protocol differences and the resulting comparison limits are summarized in
Online Resource~1.

\section{Problem Formulation and Statistical Contract}
\label{sec:problem}

Let source clients $\mathcal{S}$ train a checkpoint $f_0$ without data from
target outer unit $u$. At deployment, $u$ provides an unlabeled calibration
stream
\begin{equation}
 C_{u,b}=\{x_i\}_{i=1}^{b}, \qquad b\in\{0,32,128,512\},
\end{equation}
and an onboarding method $m$ returns
\begin{equation}
 f_{u,m,b}=A_m(f_0,C_{u,b}).
\end{equation}
The evaluation set $T_u$ belongs to recordings disjoint from calibration and is
never used by $A_m$, checkpoint selection, or hyperparameter choice. Budget
$b=32$ is the frozen primary analysis; the other budgets are supporting
sensitivity analyses.

The primary endpoint is balanced accuracy. For an HHAR user with device
endpoints $e\in\mathcal{E}_u$, metric $M$ is first averaged within seed $s$,
then across repeated seeds:
\begin{align}
 \bar M_{u,s,m,b} &= \frac{1}{|\mathcal{E}_u|}
 \sum_{e\in\mathcal{E}_u} M_{e,s,m,b}, \\
 \bar M_{u,m,b} &= \frac{1}{|\mathcal{S}_u|}
 \sum_{s\in\mathcal{S}_u}\bar M_{u,s,m,b}.
\end{align}
Other datasets have one endpoint per outer subject/user. Means, sample standard
deviations, quantiles, and intervals are then computed across $u$. Thus neither
windows, device endpoints, nor seeds increase the participant sample size.

This seed-averaged quantity estimates expected performance over the three
completed training/calibration realizations. A deployed system, however, uses
one checkpoint and one calibration sample. We therefore add a secondary stress
audit of
\begin{equation}
 G_{u,s,m,b}=100\left[\ba_u(f_{u,s,m,b})-\ba_u(f_{u,s,m,0})\right]\ \mathrm{pp},
\end{equation}
after averaging only endpoints nested within the same person and seed. The
$38\times3=114$ seed--person values per method are repeated observations of 38
people, not 114 independent participants; they receive no population-level
confidence interval. The configured seed also determines both source-training
randomness and deterministic calibration recording/window ranking, so this
audit exposes joint deployment variability but cannot decompose its two
sources.

Adaptation gain compares a method with its own zero-shot branch:
\begin{equation}
 G_{u,m,b}=100\left[\ba_u(f_{u,m,b})-\ba_u(f_{u,m,0})\right]\ \mathrm{pp}.
\end{equation}
Negative transfer is $G_{u,m,b}<0$; material negative transfer (hereafter
``material harm'') is predeclared as $G_{u,m,b}<-2$ pp. We report both rates and the worst observed gain. This
within-branch quantity must be distinguished from cross-method absolute BA:
for example, an adaptation method can repair a weak uncalibrated state yet
remain below a stronger source model.

Uncertainty uses deterministic 10,000-draw bootstrap resampling of outer units.
Paired comparisons resample participant-level differences. We emphasize effect
sizes, sample SD, intervals, and denominators rather than uncorrected
significance stars across the many method--domain--budget comparisons. With
only 6--9 outer units per headline domain, intervals are descriptive estimates,
not guarantees of population-level safety. Whenever zero material-harm events
are observed, we report the exact one-sided 95\% Clopper--Pearson upper bound
$1-0.05^{1/n}$ rather than interpreting zero observations as zero population
risk.

\subsection{Onboarding validity contract}

We treat an onboarding result as interpretable only when six conditions hold.
First, the target person is absent from source optimization and source-only
normalization. Second, calibration and evaluation derive from disjoint physical
recordings, not merely different overlapping windows. Third, every method's
source checkpoint and adaptation dependencies are identified by fold and seed.
Fourth, adaptation and checkpoint selection cannot read target evaluation
labels. Fifth, endpoints and seeds are averaged before inference across held-out
people. Sixth, unavailable calibration budgets remain missing rather than being
filled from evaluation data. These conditions define the evaluation object; a
high mean obtained after violating one of them does not answer the same
deployment question.

The contract also requires two outcomes for every adaptive branch. Absolute
post-onboarding BA measures whether the final predictor is useful relative to
other methods, while $G_{u,m,b}$ measures whether onboarding improved that
branch for person $u$. Neither outcome substitutes for the other. We therefore
do not call the method with the largest mean gain the best method, and we do not
call a method safe merely because its mean gain is positive.

\section{Leakage-Controlled Benchmark}
\label{sec:protocol}

\subsection{Datasets, preprocessing, and outer splits}

Table~\ref{tab:datasets} describes the exact local artifacts used, which can be
narrower than their parent datasets. Each artifact preserves sample, user,
domain, and recording identifiers. Normalization statistics are estimated from
source data only. Split identities and artifact hashes were frozen for the
executed protocol, and every
calibration/evaluation overlap check must pass before a run is reportable.

\begin{table}[!htbp]
\centering
\caption{Frozen local artifacts and primary outer-unit coverage. Window length
and stride are in samples; $C$ is the local class count. H denotes headline
development evidence and D descriptive evidence.}
\label{tab:datasets}
\resizebox{\textwidth}{!}{%
\begin{tabular}{llllrrl}
\toprule
Dataset/task & Local input & Win./stride & $C$ & $n$ & Tier & Unit \\
\midrule
HHAR/HAR & 6-axis acc.+gyro. & 128/64 & 6 & 9 & H & user \\
PAMAP2/HAR & chest+ankle, 12-axis & 128/64 & 9 & 8 & H & subject \\
WISDM/HAR & 3-axis accelerometer & 128/64 & 6 & 7 & H & user \\
SisFall/HAR+fall & 6-axis inertial & 200/100 & 34 & 8 & H & subject \\
UniMiB/HAR+fall & 3-axis accelerometer & 151/native & 17 & 6 & H & subject \\
UP-Fall derivative & 33-joint $xyz$ skeleton & 20/20 & binary impact & 1 & D & subject \\
\bottomrule
\end{tabular}}
\end{table}

For HHAR, device predictions are endpoints nested within the same user. The
headline unseen-user analysis covers all nine users; three leave-one-device-model
folds remain descriptive. PAMAP2 has eight eligible subjects: subject 109 was
excluded by a pre-performance feasibility rule because only one usable
recording prevented recording-disjoint calibration and evaluation. UCI HAR was
not included because the distributed window artifact did not provide auditable
trial identities. The UP-Fall input is a local five-subject skeleton derivative,
not the complete 17-subject multimodal benchmark; its single frozen outer unit
supports only a descriptive impact-phase example.

For each target endpoint, 40\% of eligible recordings form the calibration
pool under the frozen manifest and the remainder form evaluation. Requested
windows are drawn only from the calibration pool. If a client cannot supply a
budget, that outer unit is marked unavailable; windows are never borrowed from
evaluation. Labels are retained only for offline scoring. Within each
experiment seed, recordings and calibration windows are deterministically
ranked from that seed; consequently, the three seeds jointly vary source
training and calibration composition rather than acting as pure retraining
replicates.

An early HHAR engineering artifact contained index-derived channels and
overlapping physical episodes. It was invalidated before the present evidence
was frozen. None of its results is pooled here; the corrected six-axis artifact
and all broad/rotation hashes agree. This incident motivated the explicit
artifact and episode-overlap audit rather than being concealed as preprocessing.

\subsection{Source model and federated training}

The common GroupNorm backbone is a temporal convolutional network (TCN). A
pointwise input projection maps channels to width 96. Three residual
depthwise--pointwise temporal blocks use kernel 5 and dilations 1, 2, and 4,
with eight-group normalization, GELU activations, and dropout 0.1. Temporal
mean and standard deviation are concatenated into 192 features, projected to a
128-dimensional embedding, and classified by a global head. FedBN substitutes
BatchNorm for GroupNorm.

Source FL uses at most 200 rounds, 30\% client participation, one local epoch,
Adam with learning rate $5\times10^{-4}$, batch size 256, server learning rate
1, gradient clipping at 10, and update-norm clipping at 10. Class-balanced
weights are clipped to $[0.25,3]$; fall classes receive weight 2 and the fall
auxiliary loss weight is 0.30. Validation occurs every ten rounds. Early
stopping starts after round 40, uses patience four and minimum improvement
0.001, and restores the best protocol-validation checkpoint at the designated
32-window selection budget. Target evaluation labels never select a checkpoint.

\subsection{Development selection and hyperparameter freeze}

The reported values are not an untuned collection of library defaults. They
come from a staged development process: engineering screens and bounded pilots
were used to select the model width, optimization schedule, calibration
budgets, adaptation constraints, and candidate-specific settings; the corrected
multi-seed protocol then froze them before the final rotation extension. Some
training values were inherited from the already locked wearable FL recipe,
whereas method-specific values were selected on development episodes or
development summaries. Online Resource~1, Table~S3b records this provenance rather
than describing every value imprecisely as the result of the same search.

This distinction matters. Target evaluation labels are never used inside an
onboarding job, to choose its checkpoint, or to decide whether an individual
client adapts. Nevertheless, earlier development outcomes informed the final
architecture, hyperparameters, primary budget, and paper framing. The final
five-domain effect estimates are therefore development estimates. The 270-job
HHAR/PAMAP2 extension is stronger evidence against a convenient-fold artifact
because its methods, values, primary budget, and reporting rules were frozen
before those added folds were evaluated; it is not an independent confirmation
study. No post-rotation parameter was retuned.

\subsection{Core onboarding methods}

The six core methods are the only methods given the full HHAR and PAMAP2 rotations. Static controls are repeated across budgets to
make the no-adaptation reference explicit. FedBN, ATP-style, and Feature-only
consume calibration data in different ways.

The six fully rotated core strategies span three static controls and three
calibration-consuming mechanisms. FedAvg is the global GroupNorm-TCN source
control; FedPer-GH evaluates an unseen client through the global head learned
alongside private source heads; and FedPer-matched is the capacity-matched
FedPer control. FedBN-recalibration re-estimates target BatchNorm statistics
without target-trainable parameters. The ATP-style port starts from the
fold/seed-matched FedAvg checkpoint and applies one rate-scaled entropy step,
updating approximately 56--60k target parameters. Feature-only starts from the
same matched FedAvg checkpoint, freezes the network and classifier, and aligns
target embedding moments without target-trainable parameters.

FedPer source clients train private heads alongside a global head; an unseen
client has no private head and is evaluated through the global head. The
parameter-matched variant selects the nearest group-compatible backbone width
to the shared capacity of the full onboarding architecture, preventing a small
FedPer model from serving as the only personalization comparison.

FedBN keeps BatchNorm affine/statistical components local during source FL. At
onboarding, running statistics are reset and estimated without gradients from
the calibration tensor. Its zero-shot state lacks target-local statistics, so
its same-branch calibration gain is expected to be large; absolute adapted BA,
not that gain alone, is the fair cross-method endpoint.

The ATP-style port starts from the exact fold/seed-matched FedAvg checkpoint.
It learns per-parameter-tensor adaptation rates from recording-disjoint,
labeled \emph{source} episodes, then applies one entropy step to a fresh model
copy using the unlabeled target calibration stream. The source meta learning
rate is 0.01 and rates are bounded in magnitude by 0.1. It does not reproduce
ATP's image-specific custom-BN and cumulative online machinery and is therefore
named a port rather than an exact reproduction.

Feature-only also starts from the exact matched FedAvg checkpoint and freezes
the TCN and classifier. It estimates target embedding moments, shrinks them
toward source moments with strength 64, and clips per-feature scales to
$[2/3,1.5]$. The full prior-aware branch's prototypes, prior correction, and
temporal smoothing are disabled; prediction-shift safety checks remain enabled.
No target label, pseudo-label prototype, or trainable target parameter is used.

\subsection{Evidence layers and reporting outcomes}

The broad layer contains 567 jobs: 21 method labels, nine original
domain--fold instances, and seeds 42, 137, and 2026. Besides the core methods,
it includes centralized/source references, GroupNorm-compatible TENT/EATA
proxies, a head-meta proxy, a generic federated-TTA proxy, local TTA, the full
prior-aware candidate, and mechanism ablations. Proxy results are not treated
as exact external reproductions. The predeclared extension adds 270 jobs for
HHAR folds 1--8 and PAMAP2 folds 1--7 over the six core methods. The 837-job
union is therefore not a uniform $21\times24$ method--fold matrix, and the job
count is execution coverage rather than statistical sample size.

We report accuracy, macro F1, BA, p10 and worst outer-unit BA, calibration gain,
negative-transfer and material-harm rates, and calibration/adaptation/inference
time. For fall datasets, sensitivity, specificity, precision, F1, and
TP/FP/FN/TN are retained together. Online Resource~1 gives the full core
tables, method inventory, count table, protocol parameters, and artifact map.

\section{Results}
\label{sec:results}

\subsection{Coverage and evidence integrity}

All 837 declared jobs completed. The combined audit found no failed, missing,
duplicate, or overlapping run identity and verified all 18 frozen derived
files. At the primary budget, every one of the 147 method--domain rows contains
all potential outer units and all three seeds; no primary outer row has a seed
count other than three. The six core methods cover nine HHAR users and eight
PAMAP2 subjects. There are 147 non-primary outer rows with partial seed
coverage, so all higher-budget plots retain their available denominators and
remain supporting evidence. The audit status is
\texttt{passed\_with\_claim\_constraints}, not ``confirmed.''

A separate no-training reviewer-robustness package verifies all 684
primary-budget seed--person--method records ($38$ people, six methods, three
seeds), derives paired leader uncertainty and zero-event bounds, and leaves the
frozen 18-file evidence package unchanged.

The two evidence layers have different inferential roles. Complete core-method
rotations on HHAR and PAMAP2 support the split-sensitivity analysis below. The
other three headline domains corroborate cross-dataset utility and harm but do
not receive newly expanded rotations. Non-core methods retain only the original
HHAR/PAMAP2 fold and are never ranked against the rotated core means;
Online Resource~1, Table~S11 exposes every broad-screen value and denominator.

\subsection{RQ1: a single held-out person is not a stable benchmark}

Table~\ref{tab:rotation-instability} compares the original frozen fold with the
completed outer-person rotations. On HHAR, every original-fold result is above
95\% BA, while the nine-user means are 13.8--17.8 pp lower. The displayed maximum
changes from parameter-matched FedPer to FedBN. PAMAP2 is less extreme: its
displayed maximum changes from FedPer-GH to parameter-matched FedPer; five of
six methods decrease by 4.0--8.0 pp, while FedBN changes by only 0.3 pp. These are
finite-sample fold-sensitivity observations, not estimates of a universal
optimism constant or proof that arbitrary single folds are biased: the original
fold is one particular person and remains included in each full mean.

\begin{table}[!htbp]
\centering
\caption{Split sensitivity and paired uncertainty at 32 calibration windows.
Panel (a) compares the original frozen fold with completed outer-person
rotations; ranges span the six core methods. Panel (b) pairs the displayed
complete-rotation mean leader with its runner-up using a 10,000-draw paired
outer-unit bootstrap. Pair wins compares only that pair; near-max counts
per-person winning margins $\leq 1$ pp}
\label{tab:rotation-instability}
\small
\textbf{(a) Original fold versus completed rotations}\\[2pt]
\begin{tabular}{lrrrr}
\toprule
Domain & $n$ & Original BA (\%) & Complete BA (\%) & Drop (pp) \\
\midrule
HHAR & 9 & 95.6--97.2 & 78.3--83.0 & 13.8--17.8 \\
PAMAP2 & 8 & 75.5--88.4 & 75.2--81.3 & 0.3--8.0 \\
\bottomrule
\end{tabular}

\medskip
\textbf{(b) Complete-rotation leader uncertainty}\\[2pt]
\resizebox{\textwidth}{!}{%
\begin{tabular}{lrrrrr}
\toprule
Domain & Leader / runner-up & Mean BA (\%) & Difference [95\% CI] (pp) & Pair wins & Near-max \\
\midrule
HHAR & FedBN / ATP-style & 83.0/82.1 & $+0.84\ [-2.51,+5.15]$ & 4/5 & 4/9 \\
PAMAP2 & FedPer-match. / Feature-only & 81.3/80.5 & $+0.87\ [-0.09,+1.88]$ & 5/3 & 3/8 \\
\bottomrule
\end{tabular}}
\end{table}

The displayed leader changes should not be mistaken for resolved method
superiority. Panel (b) of Table~\ref{tab:rotation-instability} pairs the complete-rotation
leader with its runner-up at the participant level. Both intervals include
zero, and the nominal leader wins fewer than half of HHAR pairings. Moreover,
the gap between the largest and second-largest raw per-person score is at most
1 pp for four of nine HHAR users and three of eight PAMAP2 subjects.

\begin{figure}[!htbp]
\centering
\includegraphics[width=\textwidth]{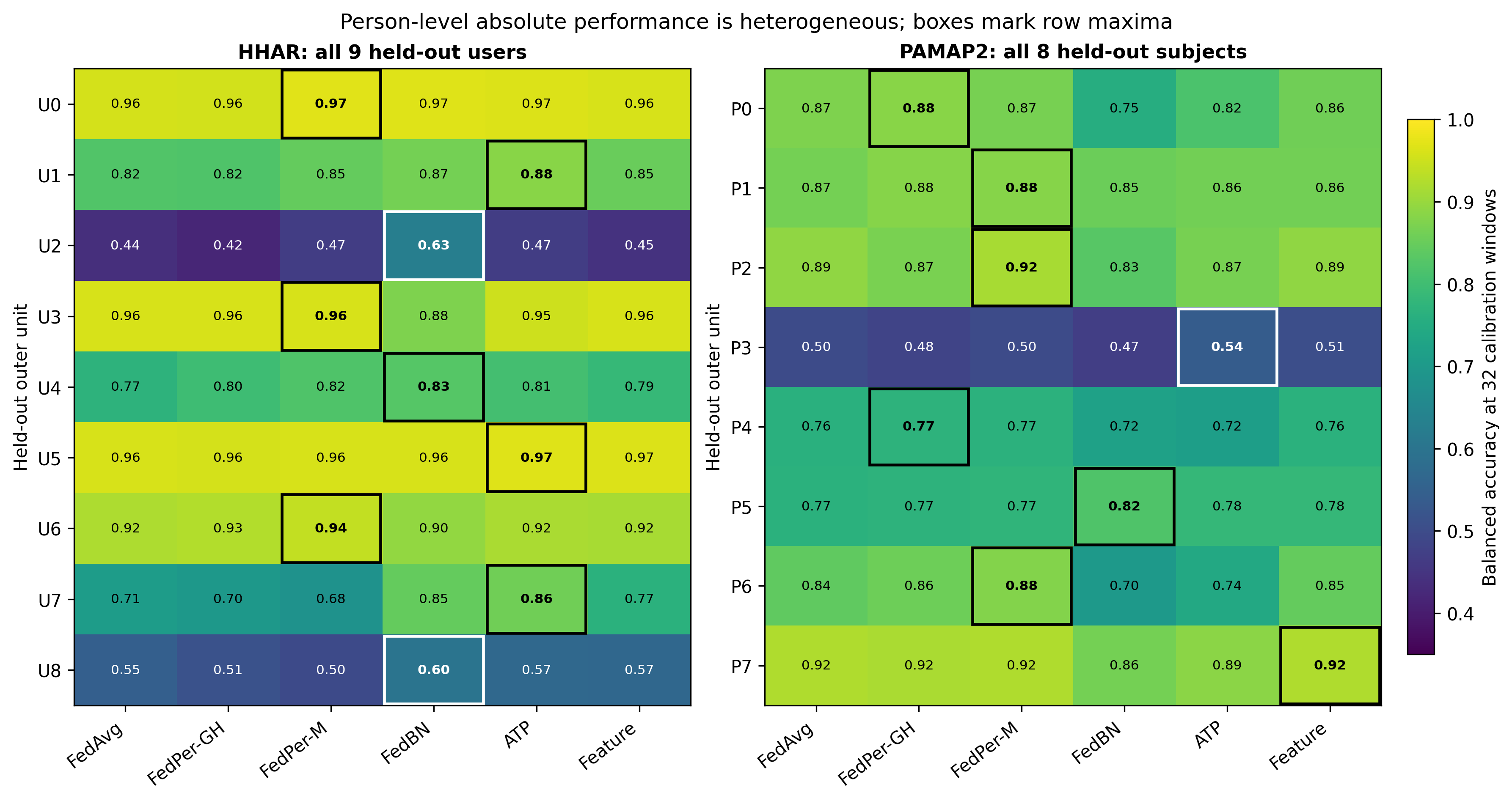}
\caption{Seed-averaged participant BA for the two complete rotations. Boxes
mark each row's numerical maximum; they do not denote statistically significant
superiority. The original fold is row U0/P0}
\label{fig:person-heatmap}
\end{figure}

Raw per-person maxima are nevertheless heterogeneous. On HHAR, ATP-style,
FedBN, and parameter-matched FedPer each supply the largest displayed score for
three of nine users. On PAMAP2, parameter-matched FedPer supplies three maxima,
FedPer-GH two, and ATP-style, FedBN, and Feature-only one each. These are
descriptive maxima, including near-ties, not statistically established
person-specific winners. Fig.~\ref{fig:person-heatmap} exposes every score and
the complete per-person margins are machine-readable.

\subsection{RQ2: absolute utility and branch-matched harm}

\subsubsection{Absolute post-onboarding quality}

Online Resource~1, Table~S4 gives complete primary-budget accuracy, macro-F1, BA, p10, and worst-unit BA; the headline BA means are summarized here. FedBN-recalibration has the
largest displayed mean on HHAR; parameter-matched FedPer leads PAMAP2, SisFall,
and UniMiB; ATP-style leads WISDM. No method leads all five datasets. The outer
SD is often large relative to between-method differences, so boldface denotes
only the maximum displayed mean and not a statistically established winner.

Baseline choice changes the apparent conclusion. Feature-only exceeds FedAvg
by $+1.46$ pp on HHAR (95\% outer bootstrap CI $[+0.45,+2.71]$ pp) and
$+2.36$ pp on WISDM ($[+0.05,+6.01]$ pp), but is essentially tied on PAMAP2 and
UniMiB and is $-0.36$ pp on SisFall. Against ATP-style it is $-1.92$ pp on HHAR,
$-3.20$ pp on UniMiB, and $-4.82$ pp on WISDM; against parameter-matched FedPer it
is $-2.17$ pp on SisFall. All four predeclared paired comparisons and intervals
are retained in Online Resource~1, Table~S5 rather than choosing only FedAvg.

\subsubsection{Within-branch gain and harm}

Table~\ref{tab:adaptive-harm} gives within-branch gains for every adaptive core
method. ATP-style shows the clearest utility--harm trade-off: it improves HHAR,
UniMiB, and WISDM on average, including $+7.18$ pp on WISDM, but loses
$-2.64$ pp on PAMAP2. Six of eight PAMAP2 subjects experience negative transfer,
five exceed the 2-pp material-harm threshold, and the worst subject loses
$9.9$ pp. On SisFall, four of eight subjects decline and two materially
decline.

Feature-only has smaller changes. Negative gains occur for 11.1--62.5\% of
outer units depending on domain, but no observed primary-budget decline crosses
$-2$ pp after seed averaging. Zero of 38 does not estimate zero population
probability: its exact one-sided 95\% upper bound is 7.6\%, and corresponding
zero-event bounds within datasets range from 28.3\% ($n=9$) to 39.3\% ($n=6$).
This is a lower-severity pattern in the observed development sample, not a
safety certificate: the sample is small, heterogeneous datasets are not one
exchangeable population, SisFall's mean gain is negative, and target labels are
unavailable to identify harmed clients online.

FedBN's large positive same-branch gains primarily measure the value of giving
an unseen client local BN statistics. They do not imply cross-method dominance:
FedBN leads absolute BA only on HHAR and remains below the best absolute mean on
the other four domains. It also materially harms one PAMAP2 subject. This is an
example of why both within-branch gain and absolute post-onboarding quality are
needed.

In Table~\ref{tab:adaptive-harm}, gain is adapted minus the same branch's
zero-shot BA. NT is the percentage of outer units with gain $<0$; MH is the
percentage with gain $<-2$ pp.

\begin{table}[!htbp]
\centering
\caption{Adaptive-method utility and harm at 32 calibration windows.}
\label{tab:adaptive-harm}
\scriptsize
\setlength{\tabcolsep}{3.3pt}
\begin{tabular}{llrrrr}
\toprule
Domain & Method & Mean gain [95\% CI] (pp) & NT & MH & Worst (pp) \\
\midrule
\multirow{3}{*}{HHAR}
 & FedBN-recal. & $+39.4\ [+35.2,+43.7]$ & 0.0\% & 0.0\% & $+30.7$ \\
 & ATP-style & $+3.4\ [+0.9,+6.7]$ & 22.2\% & 0.0\% & $-1.0$ \\
 & Feature-only & $+1.5\ [+0.5,+2.7]$ & 11.1\% & 0.0\% & $-0.5$ \\
\midrule
\multirow{3}{*}{PAMAP2}
 & FedBN-recal. & $+24.5\ [+12.2,+35.8]$ & 12.5\% & 12.5\% & $-8.2$ \\
 & ATP-style & $-2.6\ [-5.6,+0.2]$ & 75.0\% & 62.5\% & $-9.9$ \\
 & Feature-only & $+0.2\ [-0.3,+0.8]$ & 37.5\% & 0.0\% & $-1.4$ \\
\midrule
\multirow{3}{*}{SisFall}
 & FedBN-recal. & $+21.2\ [+17.9,+24.8]$ & 0.0\% & 0.0\% & $+14.3$ \\
 & ATP-style & $-0.8\ [-2.9,+0.8]$ & 50.0\% & 25.0\% & $-6.6$ \\
 & Feature-only & $-0.4\ [-0.9,+0.1]$ & 62.5\% & 0.0\% & $-1.6$ \\
\midrule
\multirow{3}{*}{UniMiB}
 & FedBN-recal. & $+43.0\ [+36.8,+51.5]$ & 0.0\% & 0.0\% & $+33.5$ \\
 & ATP-style & $+3.4\ [+0.8,+6.1]$ & 16.7\% & 0.0\% & $-0.7$ \\
 & Feature-only & $+0.2\ [-0.7,+1.1]$ & 50.0\% & 0.0\% & $-1.5$ \\
\midrule
\multirow{3}{*}{WISDM}
 & FedBN-recal. & $+35.8\ [+22.7,+48.1]$ & 0.0\% & 0.0\% & $+3.8$ \\
 & ATP-style & $+7.2\ [+2.2,+13.0]$ & 14.3\% & 14.3\% & $-2.4$ \\
 & Feature-only & $+2.4\ [+0.0,+6.0]$ & 14.3\% & 0.0\% & $-0.9$ \\
\bottomrule
\end{tabular}
\end{table}

The conclusion is not an artifact of choosing exactly 2 pp as the harm
threshold. Online Resource~1, Table~S5b recomputes counts at 1, 2, and
5 pp over the 38 headline outer units available for each adaptive
method. ATP-style has nine losses beyond 1 pp, eight beyond 2 pp, and three
beyond 5 pp. Feature-only has four losses beyond 1 pp but none beyond 2 pp;
FedBN has one PAMAP2 loss that exceeds all three thresholds. Counts are pooled
only as a descriptive audit across heterogeneous domains; domain-specific
denominators remain in Table~\ref{tab:adaptive-harm} and Online Resource~1,
Table~S5b.

Seed averaging is appropriate for the participant-level algorithm estimand but
can conceal variability relevant to one deployment. Online Resource~1, Table~S5d therefore audits the already completed seed--person realizations without
treating them as additional participants. Feature-only's 0/38 seed-averaged
count coexists with 10/114 harmful realizations affecting eight people. The
corresponding counts are 4/114 across two people for FedBN and 22/114 across 16
people for ATP-style. The worst single realization is materially lower than the
seed-averaged worst for every adaptive method. Because seed jointly changes
training and calibration composition, these values diagnose joint
reproducibility risk rather than identify its cause.

No adaptive core method therefore occupies the desired cross-domain corner of
positive mean gain everywhere and zero observed 2-pp seed-averaged
person-level harm. FedBN satisfies the mean criterion but not the harm
criterion; Feature-only satisfies the observed seed-averaged criterion but not
the mean criterion; ATP-style satisfies neither. This is a finite-sample development-benchmark statement about three mechanisms, not an impossibility theorem for
adaptation or an estimate of population safety.

A leave-one-person-out sensitivity check addresses whether one participant
drives each domain mean. ATP-style remains positive after every possible
omission on HHAR, UniMiB, and WISDM and negative after every omission on
PAMAP2 and SisFall. Feature-only remains positive after every omission on
HHAR, PAMAP2, and WISDM and negative after every omission on SisFall; its
near-zero UniMiB conclusion changes sign for one of six omissions. FedBN's mean
gain remains positive under every omission in all five domains even though its
single severe PAMAP2 loss remains operationally relevant. Online Resource~1, Table~S5c gives
the full ranges. This distinction between mean-sign stability
and individual harm is central: a stable positive mean is not a no-harm result.

\subsection{RQ3: mean, tail, and calibration quantity answer different questions}

Online Resource~1, Table~S4 identifies the descriptive leader under mean, p10,
and worst-outer-unit BA. The criterion changes the leader on PAMAP2, SisFall,
and UniMiB. On PAMAP2, ATP-style has a lower mean and frequent harm relative to
its source checkpoint, yet its final p10 and worst BA are the largest displayed;
these statements are compatible because its source and adapted client
distribution differ from the other methods. On UniMiB, parameter-matched
FedPer has the largest mean while FedBN has the strongest observed lower tail.
With only 6--9 outer units, p10 is an interpolated descriptive order statistic
with no claim to a stable population decile; worst-unit BA is reported beside it
to make that limitation visible.

\begin{figure}[!htbp]
\centering
\includegraphics[width=\textwidth]{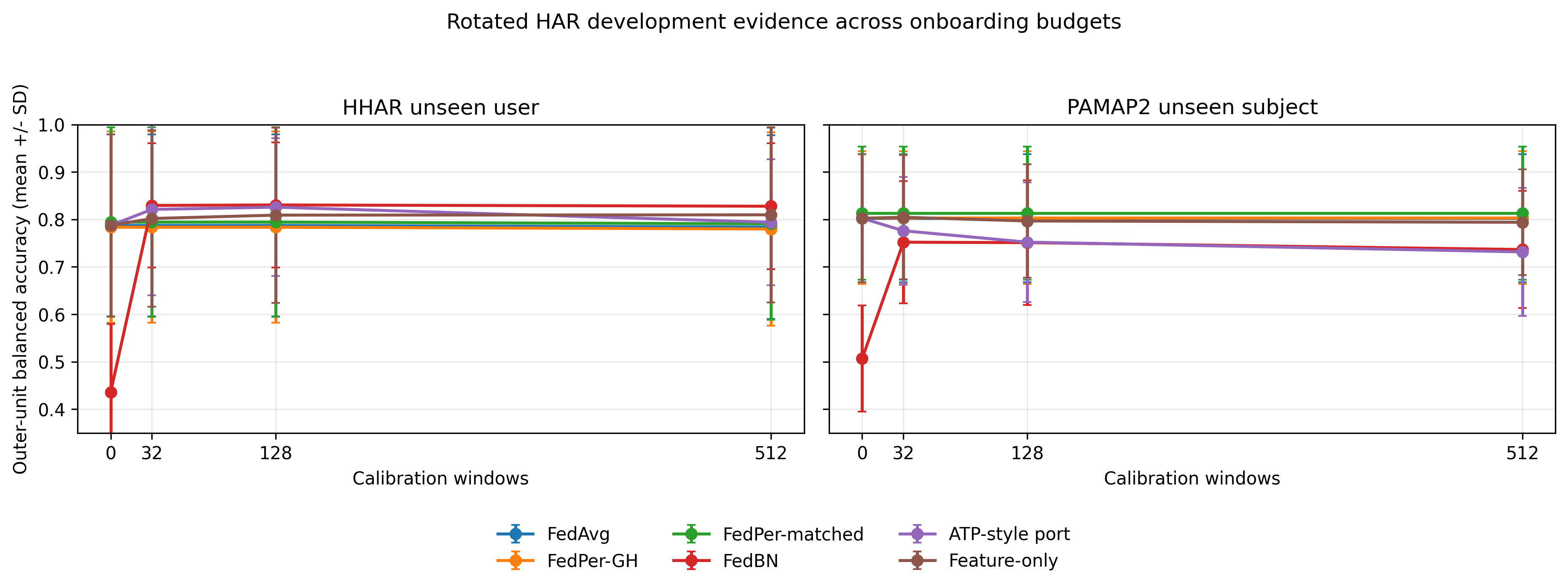}
\caption{Balanced accuracy across calibration budgets for fully rotated HHAR
and PAMAP2. Points are outer-unit means and error bars are sample SD, not
standard errors. Static controls repeat their zero-shot predictions}
\label{fig:budgets}
\end{figure}

More calibration does not establish a universal ranking either.
Fig.~\ref{fig:budgets} shows the two fully rotated HAR domains. Curves can
plateau or regress, and changes are modest compared with outer-unit SD. HHAR
and PAMAP2 are complete at the displayed budgets; on other datasets,
higher-budget availability thins and is not imputed. Budget is therefore a
resource and sensitivity axis, not an automatic safety gate.

\subsection{RQ4: fall sensitivity can conceal false-positive windows}

Table~\ref{tab:fall} reports outer-unit means, while Online Resource~1,
Table~S6 gives the corresponding pooled seed-mean TP/FP/FN/TN counts. On SisFall,
parameter-matched FedPer has the largest displayed sensitivity and F1 while
retaining 94.4\% specificity. UniMiB grouped fall/non-fall discrimination is high for several
methods even though 17-class macro F1 remains much lower (Online Resource~1,
Table~S4); the binary fall grouping is therefore not a substitute for the
multiclass endpoint. The descriptive UP-Fall derivative makes the asymmetric
error visible: Feature-only reaches 92.9\% sensitivity but 36.0\% specificity. Its
single outer unit cannot support a headline or clinical conclusion. All counts
are over evaluation windows; overlapping windows are correlated and were not
merged into alarm episodes, so neither false alarms per hour nor event-level
detection is claimed.

\begin{table}[!htbp]
\centering
\caption{Grouped fall/non-fall window performance at 32 calibration windows:
unweighted outer-unit means, reported as percentages. UP-Fall is descriptive.}
\label{tab:fall}
\small
\begin{tabular}{llrrrr}
\toprule
Domain & Method & Sens. (\%) & Spec. (\%) & Prec. (\%) & F1 (\%) \\
\midrule
\multirow{6}{*}{SisFall}
 & FedAvg & 52.9 & 94.5 & 54.9 & 40.6 \\
 & FedPer-GH & 55.3 & 93.9 & 53.8 & 41.3 \\
 & FedPer-matched & 59.4 & 94.4 & 54.4 & 43.7 \\
 & FedBN-recal. & 41.7 & 95.1 & 59.3 & 36.0 \\
 & ATP-style & 52.5 & 93.4 & 55.4 & 40.5 \\
 & Feature-only & 51.6 & 94.7 & 55.4 & 40.2 \\
\midrule
\multirow{6}{*}{UniMiB SHAR}
 & FedAvg & 99.1 & 99.6 & 99.2 & 99.2 \\
 & FedPer-GH & 98.5 & 99.6 & 99.1 & 98.8 \\
 & FedPer-matched & 97.0 & 99.3 & 98.7 & 97.8 \\
 & FedBN-recal. & 96.5 & 98.0 & 96.2 & 96.2 \\
 & ATP-style & 99.3 & 98.7 & 97.6 & 98.4 \\
 & Feature-only & 99.2 & 99.3 & 98.7 & 98.9 \\
\midrule
\multirow{6}{*}{UP-Fall (D)}
 & FedAvg & 90.7 & 39.5 & 76.2 & 82.3 \\
 & FedPer-GH & 69.8 & 68.9 & 85.4 & 74.1 \\
 & FedPer-matched & 58.1 & 64.9 & 86.9 & 60.8 \\
 & FedBN-recal. & 90.9 & 52.4 & 80.1 & 85.1 \\
 & ATP-style & 92.2 & 39.5 & 76.5 & 83.2 \\
 & Feature-only & 92.9 & 36.0 & 75.4 & 83.0 \\
\bottomrule
\end{tabular}
\end{table}

\subsection{Measured cost and negative method-selection results}

The rotation extension contains 225 source-training jobs and 45
evaluation-only Feature-only jobs. No source-training job used checkpoint
fallback. Summed per-job wall-clock measurements equal 3.51 hours on the
recorded NVIDIA L40 system; this is accounting, not suite makespan or a
hardware-independent cost claim. At 32 windows, ATP-style trains 56.1--60.0k
target parameters and takes 6.9--10.4 ms for the measured adaptation step.
Feature-only trains no parameter and takes 45.2--50.1 ms for moment estimation
and alignment. Calibration and inference times are reported separately in the
artifact. These sub-second measurements do not establish mobile energy use.

The 567-job layer also evaluates the full prior-aware candidate and its
ablations. No domain passed the frozen candidate-superiority gate. A later
source/Feature-only/ATP oracle analysis found no predeclared two-dataset
headroom sufficient to justify a deployable selector. The negative outcomes
froze the claim boundary: neither the full candidate nor a post-hoc gating
algorithm is promoted as the paper's proposed model.

\section{Discussion}
\label{sec:discussion}

\medskip\noindent\textbf{RQ1---a convenient person can change the displayed conclusion.} 
The original HHAR fold suggests near-saturation for every method and favors
parameter-matched FedPer; the complete nine-user result is much lower and has
FedBN as its largest displayed mean. PAMAP2 also changes its displayed maximum,
and five methods supply at least one raw subject-level maximum. Yet the paired
leader gaps include zero and several person-level maxima are near-ties. The
defensible result is therefore split sensitivity in absolute difficulty and
descriptive method heterogeneity, not statistically established rank reversal
or universal single-fold optimism. Complete rotation, or a prospectively
sampled participant cohort, is part of the evaluation design rather than an
optional robustness check.

\medskip\noindent\textbf{RQ2---mean utility and person-level harm define a two-dimensional
decision.} 
ATP-style obtains the largest WISDM gain and absolute score, but its PAMAP2 and
SisFall outcomes include material losses. Feature-only produces smaller gains
and no observed 2-pp case after seed averaging, while still showing four
1-pp losses, a negative SisFall mean, and harmful individual seed
realizations. FedBN improves all five dataset means but contains one
seed-averaged loss exceeding 5 pp. Threshold, zero-event-bound,
leave-one-person-out, and seed-realization analyses show that these are not
alternate wordings of one average statistic. A credible onboarding evaluation
therefore needs strong absolute baselines, a matched zero-shot branch, and
explicit harm frequency, severity, uncertainty, and realization variability.

\medskip\noindent\textbf{RQ3---neither more windows nor a positive mean is a safety test.} 
Calibration may omit classes, be dominated by one activity, or be unavailable
for short recordings. More windows can improve moment or gradient estimates,
but it does not guarantee monotonic improvement. Moreover, target labels are
absent at deployment, so the offline negative-transfer indicator cannot itself
select the correct branch. A practical abstention rule would need independent,
prospectively validated observable signals; the failed oracle headroom gate
prevents us from inventing such a selector post hoc.

\medskip\noindent\textbf{RQ4---fall-window evaluation is intrinsically asymmetric.} 
High sensitivity can coexist with low specificity and repeated false-positive
windows.
Reporting a grouped fall F1 without multiclass quality and confusion counts can
also obscure weak activity discrimination. The benchmark therefore keeps both
views and treats UP-Fall only as a descriptive impact-phase stress case. It
does not convert correlated windows into operational alarm episodes.

\medskip\noindent\textbf{What the benchmark contributes.} 
Negative benchmark results are useful when their denominators and provenance
are auditable. Here, the main scientific object is not the rejected candidate
but the onboarding validity contract: unseen-client FL, unlabeled calibration,
recording-level separation, participant-level inference, branch-matched harm,
tail quality, availability, and fall false-positive windows. The corrected HHAR history
shows why artifact-level leakage checks belong in the protocol rather than an
informal preprocessing paragraph. The rotation result adds a second warning:
even a leakage-free fold can be unrepresentative. Federated onboarding is thus
an evaluation problem before it is an optimization problem.

\section{Limitations, Validity, and Responsible Claims}
\label{sec:limitations}

\medskip\noindent\textbf{Development rather than confirmation.} 
Earlier development outcomes informed the candidate family, hyperparameters,
primary budget, and final benchmark framing. The added HHAR/PAMAP2 rotations
were predeclared after those choices were frozen, but they are still part of
the same development program rather than an independently collected
confirmation cohort. Quantitative effect estimates require independent
replication before confirmatory or safety claims. We do not report uncorrected
multiple-comparison significance tests.

\medskip\noindent\textbf{Statistical scope.} 
Headline domains contain only 6--9 outer units. Three seeds reduce stochastic
noise but do not create more participants. SisFall, UniMiB, and WISDM use the
frozen original split rather than full leave-one-person-out rotations. Only six
methods have expanded HHAR/PAMAP2 coverage; all 21 methods must not be described
as fully rotated. The leave-one-person-out sensitivity analysis evaluates
dependence on an observed participant but cannot replace a larger prospective
sample. Outer-fold source-training sets overlap substantially, so participant
bootstrap intervals do not represent uncertainty over wholly independent
training cohorts. In addition, the experiment seed jointly changes source
training and calibration composition. The seed-realization audit exposes this
variability but three joint realizations cannot separate its components.
Higher-budget missingness further limits cross-budget claims.

\medskip\noindent\textbf{Method and model scope.} 
The ATP-style port and broad-screen proxies are not author-validated exact
reproductions. FedBN uses a BatchNorm backbone while the other core methods use
GroupNorm, making its within-branch gain structurally different. The common TCN,
source-training recipe, and one-step episodic onboarding cover only one design
family. OFTTA, COA-HAR, SIGHT, PI-TTA, and other wearable-specific adaptation
methods \cite{wang2023oftta,rey2026coahar,zhou2026sight,li2026pitta} are discussed
but not executed under this protocol. The present tables
therefore compare six internal core strategies rather than claim a
state-of-the-art wearable TTA leaderboard. Different architectures or
prospective adaptation controls may change the conclusions.

\medskip\noindent\textbf{Federated scope.} 
Clients are simulated from participant partitions and only one federated source
recipe is used. The broad screen contains a centralized-source reference, but
it does not receive complete HHAR/PAMAP2 rotations and therefore cannot isolate
whether split sensitivity is unique to federated source training. The evidence
supports an evaluation protocol for checkpoints produced by this federation,
not a causal claim that federation creates the observed negative transfer.

\medskip\noindent\textbf{Dataset and deployment scope.} 
The public datasets were collected under controlled protocols, and federated
clients are simulated from participant partitions. UP-Fall is a local
five-subject skeleton derivative with one reported outer unit, not the full
multimodal benchmark. Fall events are laboratory/simulated events, so no result
establishes clinical effectiveness. Real-device latency, battery use, streaming
drift, and prospective user experience remain untested. Fall metrics are
window-level; without duration-normalized, episode-clustered alarms they do not
estimate false alarms per hour or event-level detection. The archived evidence
also does not retain every dataset's labeled calibration stream in a form that
supports a uniform post-hoc class-coverage mechanism analysis.

\medskip\noindent\textbf{Privacy, fairness, and safety.} 
Keeping raw records local in the simulation is not a differential-privacy,
secure-aggregation, or attack-resistance guarantee. No demographic fairness
claim is made; UniMiB age groups are not inferred from identifiers. ``Material
harm'' is an analysis threshold, not a medical risk threshold. We therefore
claim neither universal privacy nor deployment safety.

\section{Prospectively Scoped Future Work}
\label{sec:future}

Four experiments are deliberately left prospective rather than being simulated
through post-hoc prose. First, author code for a wearable-specific method such
as OFTTA should be ported under the same recording-disjoint outer-person
protocol and evaluated on complete HHAR/PAMAP2 rotations. Second, the frozen
analysis and harm thresholds should be tested on an untouched participant
cohort or dataset. Third, centralized and federated source checkpoints should
be compared under identical outer splits to isolate the contribution of source
federation. Fourth, each fixed training checkpoint should be evaluated under
multiple independently indexed calibration-recording/window draws, allowing
training and calibration variance to be separated. Event-clustered fall alarms,
multiple backbones, streaming drift, and real-device energy belong to the same
prospective program. None is claimed as completed evidence here, and future
results must not erase the present negative outcomes.

\section{Reproducibility and Artifact Availability}
\label{sec:reproducibility}

The current evidence package contains both protocol YAML files in the
repository, reporting scripts, job-level configurations and metrics, all
derived CSVs and figures, source-archive hashes, a hostile-review audit, and a
final SHA-256 registry. The final manifest records 567 broad jobs, 270 rotation
jobs, 837 combined jobs, 18 hashed derived files, the development role, primary
budget, statistical unit, and claim ceiling. These materials permit exact table
regeneration and coverage auditing without relying on values copied from this
PDF.

The reproducibility claim is intentionally narrower than end-to-end retraining.
The rotation archive retains per-job split provenance, but the canonical split
files referenced by the protocols are not present in the current checkout and
the broad archive does not contain an equivalent per-job provenance file.
Therefore the present package does not yet claim a from-scratch recreation of
every source split. A submission release must include the verified canonical
manifests or explicitly preserve this limitation; hashes alone are not a
substitute for their contents.

The original 18-file package remains unchanged. A first hash-registered
sensitivity package reads its frozen seed-averaged outer-unit CSV and produces
per-person maxima, harm-threshold counts, and leave-one-person-out ranges. A
second no-training package reads that registered CSV plus the archived
client-metric files, verifies 684 primary seed--person--method rows, and produces
the leader intervals, zero-event bounds, seed-realization audit, and
participant heatmap. Neither package adds a training run or participant.

\section{Conclusion}
\label{sec:conclusion}

Unlabeled onboarding of unseen federated wearable clients is conditional, not
automatically beneficial. The most consequential result is not an average
leaderboard: one frozen HHAR person yields near-saturated scores, whereas the
complete rotation lowers every mean by 13.8--17.8 pp. The displayed leader changes,
but paired intervals identify the top methods as unresolved near-ties rather
than prove a rank reversal. Across people and datasets, no adaptive core
mechanism simultaneously provides positive mean gain everywhere and zero
observed 2-pp seed-averaged harm. ATP-style combines gains with multiple
severe losses; Feature-only has lower seed-averaged severity but harmful
individual realizations and no finite-sample safety guarantee; and FedBN shows
that a stable positive mean can coexist with a severe loss and conditional
absolute quality. Calibration quantity, lower-tail outcomes, and fall-window
specificity expose further failures hidden by mean accuracy. Federated
onboarding should therefore be evaluated under an explicit validity contract:
leakage controls, complete or prospectively sampled outer-person units, matched
source checkpoints, separate participant and realization estimands, availability
denominators, and preserved negative results before deployment claims are
entertained.

\subsection*{Acknowledgement}

This work was supported by the Variable Energy Cyclotron Centre (VECC), Department of Atomic Energy (DAE), Government of India (GoI), through the provision of GPU resources and computational facilities essential for carrying out this research. The authors gratefully acknowledge the technical support and computational infrastructure provided by VECC. The authors also sincerely appreciate the peer reviewers for their insightful comments and constructive criticism, which helped improve the quality of this work.

\subsection*{Author contributions statement}

R.A. contributed to conceptualization, methodology, software development, experimental implementation, data analysis, visualization, and preparation of the original manuscript draft. V.K.R. contributed to conceptualization, methodology, experimental design, validation, interpretation of results, and manuscript writing and revision. S.M. contributed to methodology, analysis and interpretation of results, critical review, and manuscript revision. T.S. contributed to supervision, research guidance, validation, interpretation of results, and critical review and revision of the manuscript. All authors reviewed and approved the final manuscript.

\subsection*{Additional information}

\textbf{Accession codes}: Not applicable.\\
\textbf{Competing interests}: All authors assert that they own no financial or personal affiliations that may be seen as affecting the work provided in this study. No conflicts of interest are acknowledged.

\subsection*{Declarations}

\subsubsection*{Data availability}
All source code, configuration files, and supplementary scripts used in this study are publicly available at \url{https://github.com/TitanRahil/When-Adaptation-Hurts}. The datasets are available under its original license and cannot be redistributed by the authors. Due to institutional data-sharing restrictions, detailed training logs and key result files are not openly available but can be accessed upon reasonable request to the corresponding author at reachme@soumyamazumdar.com.


\subsubsection*{Clinical Trial Number}
Clinical trial number: not applicable.

\subsubsection*{Ethics, Consent to Participate, and Consent to Publish}
Ethics, Consent to Participate, and Consent to Publish declarations: not applicable.

\subsubsection*{Conflict of Interest}
All authors assert that they own no financial or personal affiliations that may be seen as affecting the work provided in this study. No conflicts of interest are acknowledged.

\bibliographystyle{IEEEtran}
\bibliography{privfedtalk_refs}

\section*{Author Biographies}

\renewcommand{\arraystretch}{1.2}
\noindent\begin{tabular}{@{}p{0.17\textwidth} p{0.78\textwidth}@{}}

\begin{minipage}[t]{\linewidth}
\vspace{0pt}
\includegraphics[width=\linewidth]{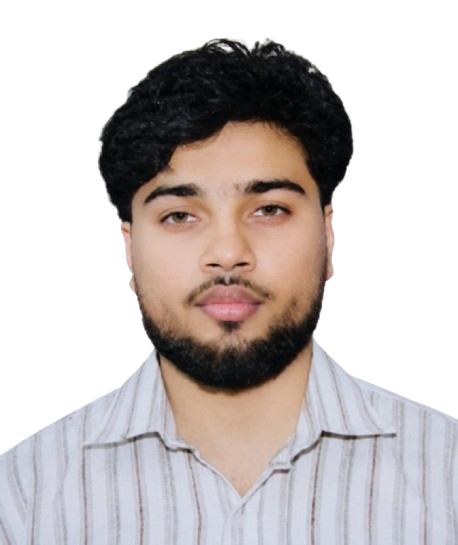}
\end{minipage}
&
\begin{minipage}[t]{\linewidth}
\vspace{0pt}
\textbf{Rahil Aftab} is an undergraduate student pursuing a Bachelor of Technology (B.Tech.) in Computer Science and Engineering at Jamia Hamdard, New Delhi, India. His research interests include artificial intelligence, computer vision, federated learning, privacy-preserving machine learning, and large language models. He has contributed to research projects involving federated learning, benchmark evaluation, and AI-driven applications. He completed a research internship at the Variable Energy Cyclotron Centre (VECC), Department of Atomic Energy, India, where he conducted research on federated learning for privacy-preserving AI systems. His broader interests include trustworthy AI and the real-world deployment of intelligent systems.
\end{minipage}
\\[1.5em]

\begin{minipage}[t]{\linewidth}
\vspace{0pt}
\includegraphics[width=\linewidth]{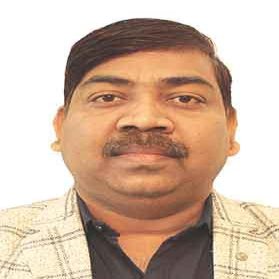}
\end{minipage}
&
\begin{minipage}[t]{\linewidth}
\vspace{0pt}
\textbf{Vineet Kumar Rakesh} is a Technical Officer (Scientific Category) at the Variable Energy Cyclotron Centre (VECC), Department of Atomic Energy, India, with over 23 years of experience in software engineering, database systems, and artificial intelligence. His research focuses on talking head generation, lip reading, and ultra-low-bitrate video compression for real-time teleconferencing. He is currently pursuing a Ph.D. at Homi Bhabha National Institute, Mumbai. Mr. Rakesh has contributed to office automation, OCR systems, and digital transformation projects at VECC. He is an Associate Member of the Institution of Engineers (India) and a recipient of the DAE Group Achievement Award.
\end{minipage}
\\[1.5em]

\begin{minipage}[t]{\linewidth}
\vspace{0pt}
\includegraphics[width=\linewidth]{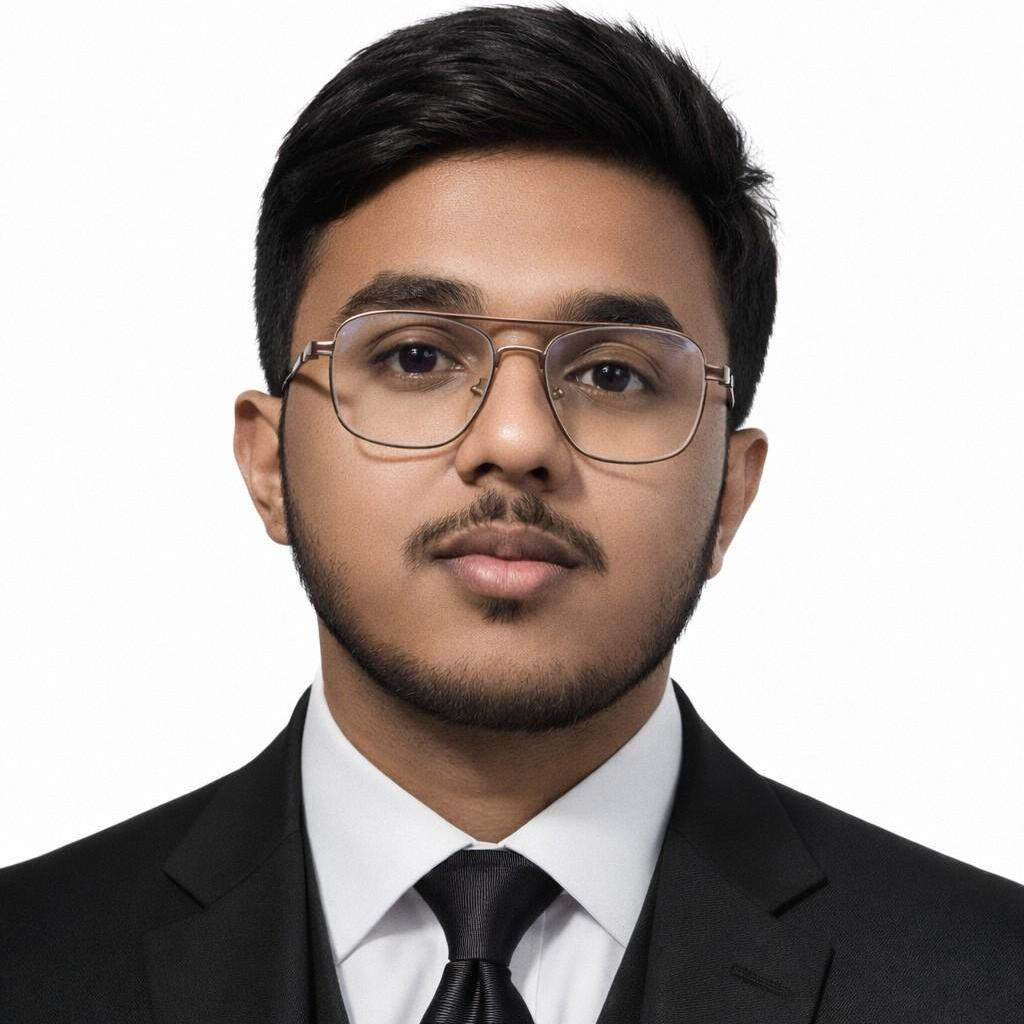}
\end{minipage}
&
\begin{minipage}[t]{\linewidth}
\vspace{0pt}
\textbf{Soumya Mazumdar} is a student researcher pursuing a B.S. in Data Science and Applications at the Indian Institute of Technology Madras and B.Tech. in Computer Science and Business Systems at West Bengal University of Technology (GMIT campus), India. His work focuses on temporal generative modeling, geometry-aware computer vision, and controllable video synthesis, with particular interest in diffusion-based methods for talking-head generation and temporal consistency. He has served as a Research Trainee at the Variable Energy Cyclotron Centre (VECC), where he worked on pose- and landmark-conditioned video generation, benchmarking, and efficient deployment pipelines. He has contributed to research publications in journals, conference proceedings, and edited volumes, and is also associated with an Indian patent in neural network-based real-time analysis.
\end{minipage}
\\[1.5em]

\begin{minipage}[t]{\linewidth}
\vspace{0pt}
\includegraphics[width=\linewidth]{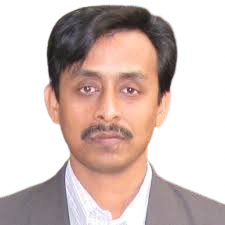}
\end{minipage}
&
\begin{minipage}[t]{\linewidth}
\vspace{0pt}
\textbf{Dr. Tapas Samanta} is a senior scientist and Head of the Computer and Informatics Group at the Variable Energy Cyclotron Centre (VECC), Department of Atomic Energy, India. With over two decades of experience, his work spans artificial intelligence, industrial automation, embedded systems, high-performance computing, and accelerator control systems. He also leads technology transfer initiatives and public scientific outreach at VECC.
\end{minipage}
\\

\end{tabular}
\end{document}